\documentclass[lettersize,journal]{IEEEtran}
\usepackage{amsmath,amsfonts}
\usepackage{algorithmic}
\usepackage{algorithm}
\usepackage{array}
\usepackage[caption=false,font=normalsize,labelfont=sf,textfont=sf]{subfig}
\usepackage{textcomp}
\usepackage{stfloats}
\usepackage{url}
\usepackage{verbatim}
\usepackage{graphicx}
\usepackage{cite}
\usepackage{algorithm}
\usepackage{algorithmic}
\usepackage{multirow} 
\usepackage{amssymb}
\usepackage{booktabs}
\usepackage{hyperref} 
\usepackage[dvipsnames]{xcolor}
\begin{document}

\title{DiffMOD: Progressive Diffusion Point Denoising for Moving Object Detection in Remote Sensing}


\author{
    \IEEEauthorblockN{Jinyue Zhang\IEEEauthorrefmark{2},
    Xiangrong Zhang\thanks{* Xiangrong Zhang is the corresponding author.}\IEEEauthorrefmark{2}\IEEEauthorrefmark{1}, Senior Member, IEEE, 
    Zhongjian Huang\IEEEauthorrefmark{2}, 
    Tianyang Zhang\IEEEauthorrefmark{2},
    Yifei Jiang\IEEEauthorrefmark{3}, and
    Licehng Jiao\IEEEauthorrefmark{2}, Fellow, IEEE}
    
    \IEEEauthorblockA{
        \IEEEauthorrefmark{2}Xidian University, China, \href{mailto:xxx@xxx,xxx@xxx}{xrzhang@mail.xidian.edu.cn}\\
        \IEEEauthorrefmark{3}Inspur Software Co., Ltd., China,
        }
}



\maketitle

\begin{abstract}
    Moving object detection (MOD) in remote sensing is significantly challenged by low resolution, extremely small object sizes, and complex noise interference. 
    Current deep learning-based MOD methods rely on probability density estimation, which restricts flexible information interaction between objects and across temporal frames. 
    To flexibly capture high-order inter-object and temporal relationships, 
    we propose a point-based MOD in remote sensing. 
    Inspired by diffusion models, the network optimization is formulated as a progressive denoising process 
    that iteratively recovers moving object centers from sparse noisy points.
    Specifically, we sample scattered features from the backbone outputs as atomic units for subsequent processing, 
    while global feature embeddings are aggregated to compensate for the limited coverage of sparse point features.
    By modeling spatial relative positions and semantic affinities, 
    Spatial Relation Aggregation Attention is designed to enable high-order interactions among point-level features for enhanced object representation.
    To enhance temporal consistency,
    the Temporal Propagation and Global Fusion module is designed, which leverages an implicit memory reasoning mechanism for robust cross-frame feature integration. 
    To align with the progressive denoising process, we propose a progressive MinK optimal transport assignment strategy that establishes specialized learning objectives at each denoising level.
    Additionally, we introduce a missing loss function to counteract the clustering tendency of denoised points around salient objects.
    Experiments on the RsData remote sensing MOD dataset show that our MOD method based on scattered point denoising can more effectively explore potential relationships between sparse moving objects and improve the detection capability and temporal consistency.
\end{abstract}


\section{Introduction}

\IEEEPARstart{M}{oving} object detection (MOD) in remote sensing involves identifying and localizing moving objects of interest from high-resolution video data acquired by remote sensing satellites \cite{viso, DSFNet}. 
This technology serves as the foundation for tasks such as object tracking \cite{GAMO}, density estimation\cite{VP}, and traffic prediction \cite{Bidirectional},
which plays a significant role in fields such as environmental monitoring, urban planning, and emergency response  \cite{Brain}. 
Compared to MOD in natural scenes, satellite videos often have low resolution, wide field of view, and smaller-sized objects of interest (e.g., vehicles), resulting in limited appearance features \cite{10292941,10459254}. 
In addition, factors such as video acquisition angles, climate conditions, and lighting variations can introduce noise into video sequences \cite{10163641}.
These limitations make MOD in remote sensing particularly challenging.

\begin{figure}
    \centering
    \includegraphics[width=\linewidth]{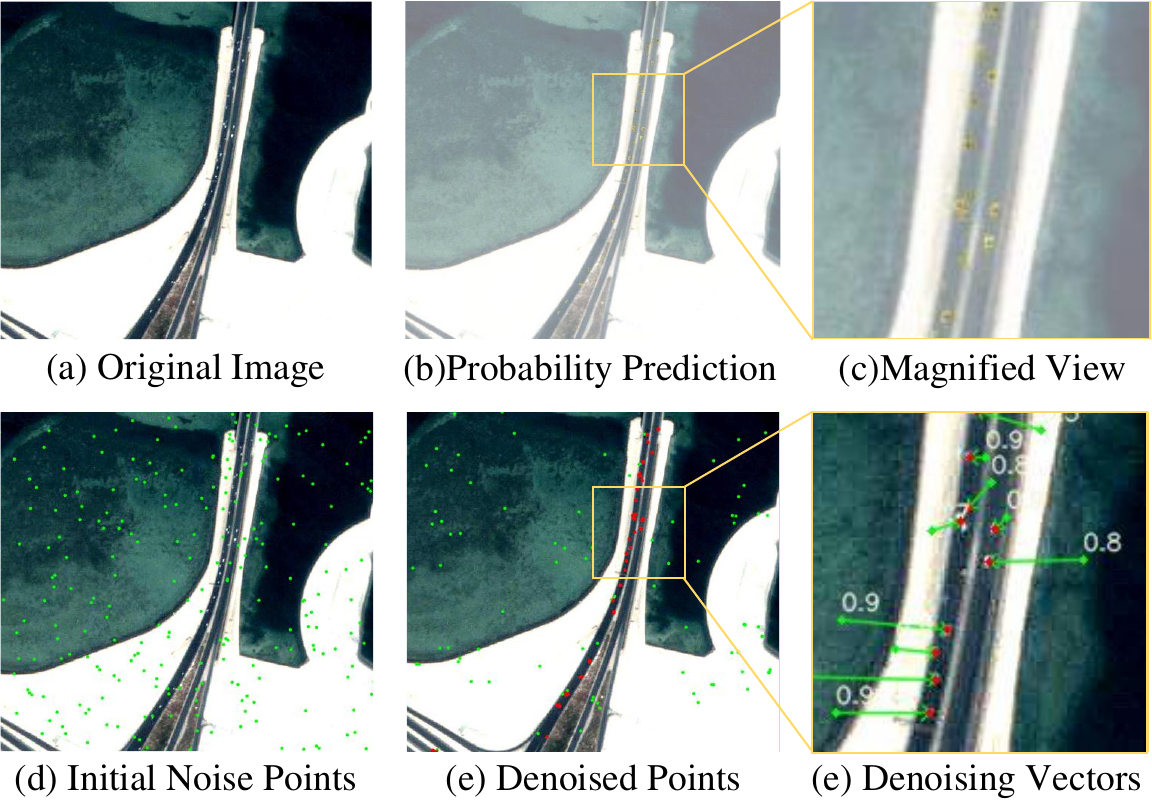}
    \caption{Comparison of existing methods (the first row) and DiffMOD (the second row).}
    \label{fig:points}
\end{figure}

Existing MOD methods in remote sensing can be categorized into two main approaches: model-based and learning-based methods. 
Classical model-based MOD \cite{E-LSD,Decolor,AGMM,B-MCMD,GoDec, ViBe} is noise-sensitive and computationally expensive.
The second category comprises learning-based methods. 
These methods \cite{Clusternet, DSFNet, GA-PANet} leverage temporal difference maps or 3D convolutions to capture short-term motion cues.
However, since moving objects of interest in remote sensing are typically small in size, 
these learning-based methods \cite{SDANet, MICPL} often perform inference and discrimination on high-resolution low-level feature maps to ensure adequate distinguishability. 
As shown in Fig.~\ref{fig:points}(b) and (c), dense probability estimation is confined to low-order local interactions through convolution operations. This approach fundamentally limits interactions of high-order information between distant objects.

Addressing the above problems, we propose a point-based MOD in remote sensing, which employs point features as atomic units to facilitate high-order relational reasoning. 
As shown in Fig.~\ref{fig:points}(d) and (e), the start inputs are sparse noise-corrupted scatter points and gradually
refined to approximate the centers of the true objects. Fig.~\ref{fig:points}(f) shows the estimated denoising vectors.
On the other hand, error accumulation during the propagation of temporal information remains a fundamental challenge in the detection and tracking of moving objects \cite{10163641, 10397084}. 
However, our progressive diffusion denoising detector fundamentally differs by modeling the noise distribution itself.
The optimization process begins with initial sparse noise points and iteratively converges toward precise target centroids through iterative denoising steps.
Since target motion is physically constrained by velocity limits \cite{9785979}, the temporally accumulated noise typically remains within permissible bounds, thus avoiding performance degradation.

In summary, a novel progressive diffusion point denoising framework is proposed for MOD in remote sensing, 
which iteratively recovers moving objects from sparse noise-corrupted scatter points.
Specifically, we sample scattered features from the dense backbone outputs as atomic units for subsequent processing, 
while aggregating global feature embeddings captured by non-overlapping sliding windows to compensate for sparse point features' limited coverage. 
Spatial Relation Aggregation Attention (SRAA) is designed to dynamically integrate object information by jointly encoding spatial relative positions and semantic affinities between features. 
It has two self-attention and cross-attention variants, named SRSA and SRCA.
In each denoising level, SRCA are used for integrating point-level features and global information.
And SRSA captures high-order interactions within point-level features.   
The denoising process operates progressively, with each level utilizing the output of the preceding level as its noisy input.
For temporal-aware modeling, the Temporal Propagation and Global Fusion (TPGF) module transforms preceding frames' scattered information into regional representations, which are memorized and propagated to dynamically adjust current-frame global features. 
During training, we introduce a progressive minK optimal transport assignment (MinK OTA) strategy that establishes level-specific learning objectives across denoising levels.
This is complemented by a target missing loss, both designed to mitigate clustering artifacts where denoised points over-concentrate around salient objects.

\begin{itemize}
\item 
    We propose a progressive diffusion point denoising framework for MOD in remote sensing (DiffMOD), 
    which employs point features as atomic units to enable high-order interactions both spatially and temporally.

\item 
    A novel attention module SRAA is designed to aggregate features by jointly modeling both relative positional relationships and semantic affinities.

\item 
    TPGF module is present to enhance temporal consistency, which leverages an implicit memory reasoning mechanism for robust cross-frame feature integration. 
\item 
To fundamentally address denoised point clustering in diffusion-based MOD, we combine progressive MinK OTA strategy with constraint enforcement through target missing loss function. 
\end{itemize}

\section{Related Work}
\subsection{MOD in Remote Sensing}
Compared to natural images, videos captured by satellites typically exhibit lower resolution, wider scenes, smaller object sizes, and more complex backgrounds. 
Relying solely on appearance and texture information makes it challenging to achieve satisfactory performance. 
Therefore, both model-based methods and learning-based methods for MOD in satellite videos are exploring more effective ways to leverage temporal information.

Model-based methods model the scene as a stable background, sparse targets, and noise.
Based on this modeling, a straightforward MOD approach involves detecting moving objects through frame differencing \cite{MMB}, 
or modeling the background \cite{ViBe, DBP} using mean or median filtering. 
However, these methods are sensitive to noise.
To address this limitation, more advanced techniques \cite{E-LSD,Decolor,AGMM,B-MCMD,GoDec}, employ low-rank and sparse decomposition, which achieves global optimization and better distinguishes objects from noise.
Although these methods have theoretical completeness,
their high computational complexity makes them unable to meet the demands of large-scale satellite video data processing.
With the remarkable advancements of deep learning in the realm of computer vision, learning-based methods for MOD in satellite videos have emerged \cite{Clusternet, MICPL, DSFNet, GA-PANet, SDANet, 10459254}. 
Clusternet \cite{Clusternet} proposes a two-stage detector that first locates potential regions in wide-area scenes, then detects small moving targets within them by combining appearance and motion cues.
DSFNet \cite{DSFNet} propose a two-stream detection network, which is composed of a 2-D backbone to extract static context information from a single frame and a lightweight 3-D backbone to extract dynamic motion cues from consecutive frames.
GA-PANet \cite{GA-PANet} designs a historical frame differential module to extract the motion information and fuse with the current frame to obtain the spatio-temporal feature. 

\begin{figure*}
    \centering
    \includegraphics[width=\linewidth]{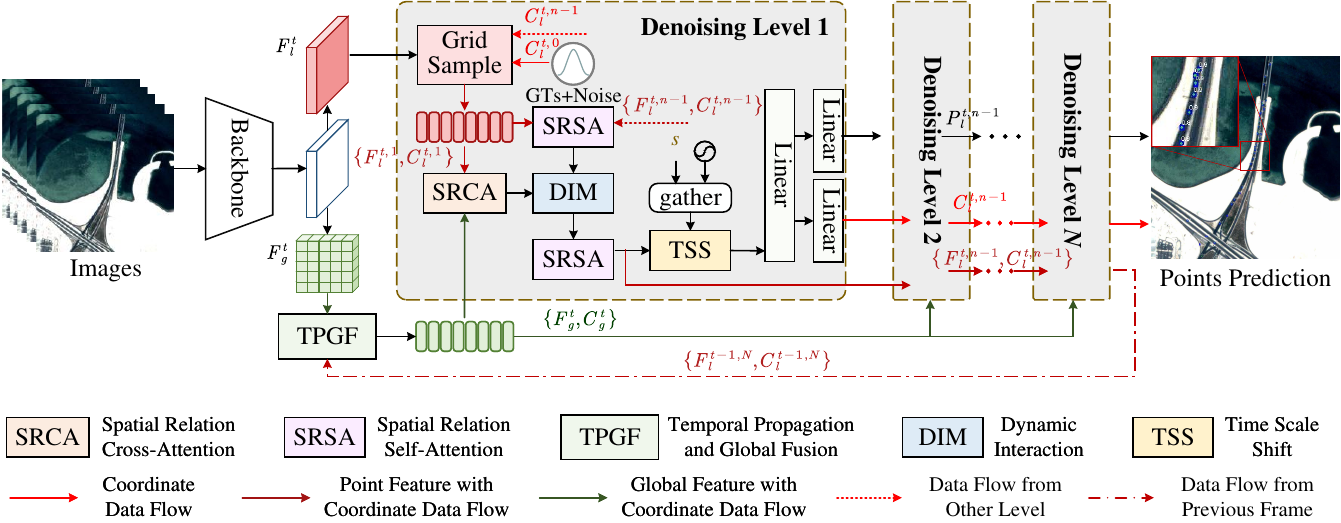}
    \caption{The progressive point denoising framework of the DiffMOD. } 
    \label{fig:structure}
\end{figure*}

\subsection{Diffusion Model}
Diffusion models have emerged as a powerful paradigm for generative modeling \cite{LatentDiff}. 
The model operates by gradually denoising the data through a Markov chain that reverses a gradual noising process. 
The noising process can be formed as:
\begin{equation}
    q\left( x_s|x_{s-1} \right) =N\left( x_{\mathrm{s}}; \sqrt{1-\beta_s} x_{s-1},\beta_s I \right) 
\end{equation}
where $0<\left\{ \beta _s \right\} _{s=1}^S<1$ is the noise factor‌, $s$ is the timestep. 
$x_s$ is the noisy data at step $s$.

The forward process systematically degrades the input signal to isotropic Gaussian noise $x_S \sim N\left( 0,I\right)$ across $S$ diffusion steps, 
whereas the reverse process employs a learned neural network to iteratively denoise and reconstruct the original data distribution.
Using the reparameterization trick, $x_s$ can be sampled directly from the initial data $x_0$ as:
\begin{equation}
    x_{\mathrm{s}}=\sqrt{\overline{\alpha }_s}x_0+\sqrt{1-\overline{\alpha }_s}\epsilon ,\epsilon \sim N\left( 0,I \right) 
\end{equation}
where $\alpha_s=1-\beta_s$ and $\overline{\alpha }_s=\prod_{s=1}^S{\alpha _{\mathrm{s}}}$

The neural network's learning objective is to minimize the mean squared error between the predicted noise and the actual noise.
\begin{equation}
L_{denoise}=\mathbb{E} _{s,x_0,\epsilon}\left[ \left\| \epsilon _{\theta}\left( x_{\mathrm{s}},s \right) -\epsilon \right\| ^2 \right] 
\end{equation}
where $\epsilon _{\theta}$ is the noise predicted by the network parameterized by $\theta$ at diffusion step $s$ and $\epsilon$ 
is the true Gaussian noise added during the forward process.
The single-step denoising process from $x_s$ to $x_{s-1}$ is formulated as:
\begin{equation}
    p_{\theta}\left( x_{\mathrm{s}-1}|x_{\mathrm{s}} \right) =N\left( x_{\mathrm{s}-1};\mu _{\theta}\left( x_{\mathrm{s}},s \right) ,\Sigma _{\theta}\left( x_{\mathrm{s}},s \right) \right) 
\end{equation}
\begin{equation}
    \mu _{\theta}\left( x_{\mathrm{s}},s \right) =\frac{1}{\sqrt{\overline{\alpha} _s}}\left( x_{\mathrm{s}}-\frac{\beta _s}{\sqrt{1-\overline{\alpha} _s}}\epsilon _{\theta}\left( x_{\mathrm{s}},s \right) \right) 
\end{equation}
where $\mu _{\theta}\left( x_{\mathrm{s}},s \right)$ is the predicted mean of the denoised distribution.


The ongoing theoretical developments and refinements in diffusion models have led to their expanding applications in discriminative tasks including, but not limited to, object detection \cite{DiffusionDet,crowddiff}, tracking \cite{diffmot,difftrack}, and trajectory prediction \cite{Diff_rntraj,IAdifftraj}.
DiffusionDet \cite{DiffusionDet} pioneered their use for object detection by framing detection as a denoising process from noisy boxes to ground-truth. 
This approach treats object queries as noisy instances in a continuous space, progressively refined through diffusion steps.
Our method is inspired by DiffusionDet \cite{DiffusionDet}, but addresses the challenges of MOD in remote sensing, where large scenes contain extremely small objects. 
Therefore, we further simplify the object proposal-based modeling in DiffusionDet into scatter point modeling. 
The saved computational resources are reallocated to higher-order relationship modeling among the scatter points.

\section{Method}
\subsection{Overview}
In this section, we introduce the progressive diffusion point denoising framework of DiffMOD in remote sensing. 
As illustrated in Fig.~\ref{fig:structure}, the framework of the diffusion-based MOD network is depicted. 
We formulate the MOD problem in remote sensing as a progressive sparse point denoising process, 
where noisy sparse points are gradually refined to approximate the centers of the true objects. 
Similarly to DSFNet \cite{DSFNet}, we employ a dual-branch spatiotemporal network as the backbone to extract dense feature representations.
Then, subsequent operations will be performed based on sparse point features.

The overall framework follows a progressive denoising paradigm. 
Within each denoising level, we extract point features $F_l^n \in \mathbb{R}^{L_p \times d}$ from the dense feature representation through grid sampling at the coordinates of noisy scatter points. 
$L_p$ is the number of noisy scatter points and $d$ is the feature dimension. 
The initial point set consists of sparse samples at Denoising Level 0, randomly distributed across the scene space. 
As part of our training protocol, we artificially introduce perturbed instances of ground-truth object centers to enhance learning.
The point set at the denoising level $n$ is derived from the output of the previous denoising level $n-1$. 
To obtain comprehensive and stable scene representations, we employ the patch embedding strategy of transformer to extract global features $F_g \in \mathbb{R}^{L_g \times d}$. $ L_g = (H/r_g)\times(W/r_g)$ is the number of patch embedding, $H,W$ is the image height and width, $r_g$ is the stride of patch embedding.

SRAA is our novel attention module that aggregates features by jointly modeling both relative positional relationships and semantic affinities.
SRCA and SRAA represent cross-attention and self-attention variants, respectively, designed for cross-feature integration and intra-feature refinement.
Here, SRCA enables the sparse point features to acquire stable scene context from global embeddings, thereby compensating for the information loss caused by their high sparsity. 
The Dynamic Interaction Module (DIM) adaptively estimates the mapping parameters to independently adjust each point feature $F_l^n$ at the current level $n$ based on features $F_l^{n-1}$ from the previous denoising level $n-1$.
Similarly to diffusion models, we employ a timestep $s$ to control noise levels during noise generation. 
The Time-Step Scaling (TSS) module, same as in DiffusionDet \cite{DiffusionDet}, adaptively scales features by acquiring embeddings corresponding to discrete time steps, thereby facilitating the progressive denoising process.
Finally, linear layers jointly predict the probability $P_l^{n}$ of each sparse point belonging to the moving objects and its denoised coordinates $C_l^{n}$.

To enhance temporal consistency in moving object detection, we design a TPGF module that distributes sparse point features $F_l^{t-1}$ from the preceding frame $t-1$ into spatially partitioned global regions according to point locations $C_l^{t-1}$, represents each region's historical information using averaged point features, and dynamically adjusts global features $F_g^{t}$ of the current frame via a GRU-based implicit memory mechanism, which encodes and propagates temporal states across sequential frames.

\subsection{Spatial Relation Aggregation Attention}
Unlike the fixed token-based information retrieval and interaction in image transformer based networks, 
the positions of sparse point features are inherently random. 
Additionally, the number of objects in a scene is unknown, 
resulting in no fixed ratio between the number of sampled sparse points and the number of objects. 
Therefore, SRAA is designed to dynamically integrate information by jointly encoding spatial relative positions and semantic affinities between features.

\begin{figure}
    \centering
    \includegraphics[width=1\linewidth]{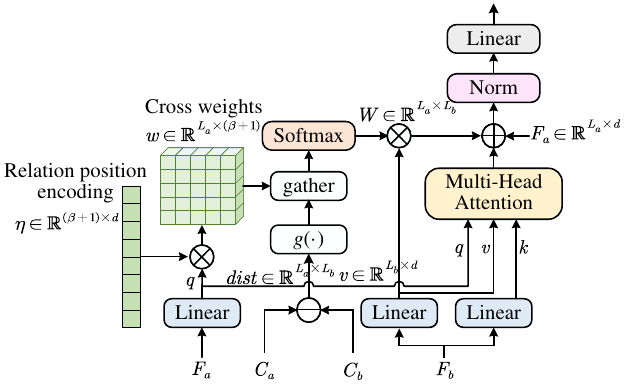}
    \caption{Spatial Relation Aggregation Attention}
    \label{fig:SRAA}
\end{figure} 

\begin{figure}
    \centering
    \includegraphics[width=0.95\linewidth]{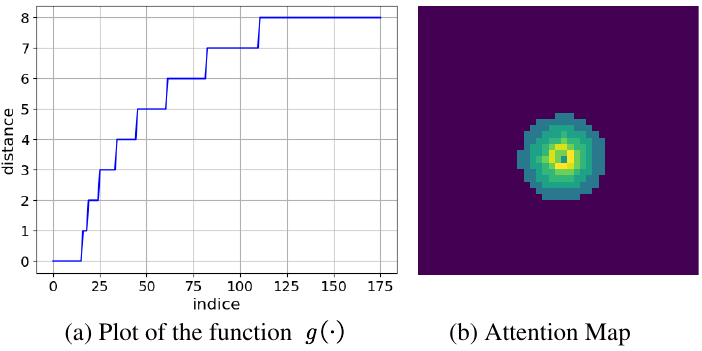}
    \caption{Function $g(x)$ and SRAA attention visualization.}
    \label{fig:gx}
\end{figure}

The structure of SRAA is illustrated in Fig.~\ref{fig:SRAA}. 
The inputs include features $f_a \in \mathbb{R}^{L_a \times d}$, features $f_b\in \mathbb{R}^{L_y \times d}$, 
and their corresponding positions $C_a\in \mathbb{R}^{L_a \times 2}$ and $C_b \in \mathbb{R}^{L_b \times d}$. 
$d$ is the is the feature dimension.
First, the distance $dist \in \mathbb{R}^{L_a \times L_b}$ between the positions is calculated. 
According to common intuition, for a given point, the influence of nearby points should be greater, and this influence should gradually decay as the distance increases. 
Therefore, we define a piecewise quantization function $g(x)$:
\begin{equation}
g\left( x \right) =\left\{ \begin{matrix}
	\lfloor \frac{x}{\alpha} \rfloor&		if\,\,x \leqslant \alpha \\
	\min \left( \beta ,\lfloor 1+\small{\frac{\ln \left( x/\alpha \right)}{\ln \gamma}\left( \beta -1 \right)} \rfloor \right)&		if\,\,x>\alpha\\
\end{matrix} \right. 
\end{equation}
Here, $\alpha$ is the stride of the spatial partitioning and is set to 16, same as $r_g$, and $\gamma$ controls the slope of the function and is set to 8. 
The Fig.~\ref{fig:gx}(a) show the plot of the function $g(x)$. 
The quantized indices comprise integer values within the range $[0, \beta]$, resulting in a cardinality of $(\beta+1)$ distinct levels, here $\beta=8$.

On the other hand, based on the semantic information of point features, the point features perform matrix multiplication with relation position encoding $\eta \in \mathbb{R}^{(\beta+1) \times d}$ to compute similarity as cross weights $w\in \mathbb{R}^{L_a \times (\beta+1)}$. 
Relation position encoding is a set of learnable vectors that interact with the query embedding.
Then, based on the index values, the spatial relational attention weights $W\in \mathbb{R}^{L_a \times L_b}$ of $C_a$ relative to $C_b$ are gathered from the cross weights and used to retrieve information from $F_b$. 
The attention weights are multiplied with the values after applying the Softmax operation, and the result is added to the output of the multi-head attention and the original input $F_a$. 
Finally, the output is obtained through normalization and a linear layer mapping.
As shown in Fig.~\ref{fig:gx}(b), the attention weight between feature $f_a$ and feature $f_b$ is stronger when their spatial distance is smaller, 
while being adaptively adjusted based on the semantic feature $f_a$ itself.

\subsection{Temporal Propagation and Global Fusion}

\begin{figure}
    \centering
    \includegraphics[width=0.9\linewidth]{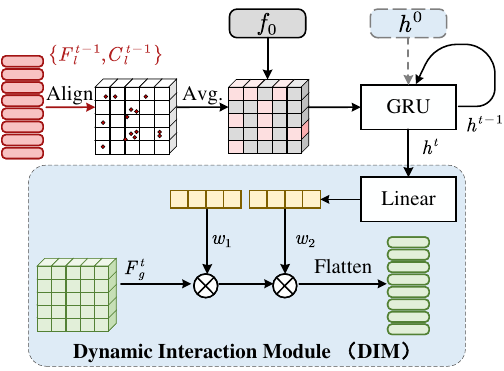}
    \caption{Temporal Propagation and Global Fusion Module}
    \label{fig:TPGF}
\end{figure}

As illustrated in Fig.~\ref{fig:TPGF}, we propose the TPGF module to enhance temporal consistency for MOD in remote sensing.
TPGF module processes sparse point features $F_l^{t-1}$ with their corresponding coordinates $C_l^{t-1}$ from the previous frame's output. 
Initially, it spatially aligns these points to predefined global regions based on their coordinates. 
For each region, the features of contained points are aggregated through averaging, while regions without any points are initialized with a learnable parameter $f_0 \in \mathbb{R}^{1 \times d}$.

The integrated features are treated as the state representations of each global region from the previous frame $t-1$. 
These region states are then combined with the preceding hidden state $h^{t-1}$ through a GRU module to predict the current frame's hidden state $h^t$, thereby enabling temporal state propagation across consecutive frames.

DIM adaptively estimates mapping parameters to independently adjust each global region feature.
The DIM utilizes the current frame's hidden state $h^t$ to estimate mapping parameters $\omega$, which dynamically adjust the features $F_g^t$ of each global region through adaptive transformation.
The mapping operation in DIM can be formally represented as:
\begin{equation}
    F_{g}^{t\prime} =\left( F_{g}^{t}*\omega _1 \right) *\omega _2
\end{equation}
where $F_{g}^{t} \in \mathbb{R} ^{L_g \times 1 \times d}$ represents original global features at frame $t$, $L_g$ is the number of global embeddings. 
The transformation matrices $\omega _1 \in \mathbb{R} ^{L_g \times d \times k}$ and $\omega _2 \in \mathbb{R} ^{L_g \times k \times d}$ split from $\omega \in \mathbb{R} ^{L_g \times 2kd}$ 
first project the features into a higher $k$-dimensional space via matrix multiplication $*$ and then mapping them back to the original $d$-dimensional space, 
thereby enabling adaptive feature adjustment.

\subsection{Training Pipeline}
During training for real-world inference simulation, the sparse point set is constructed through the following methodology: 
Ground-truth centers are corrupted with noise while incorporating uniformly distributed background scatter points. 
The noise range is progressively adjusted according to the radius parameter $r$.
\begin{equation}
    noise\sim U\left( -\sigma_s \cdot r\cdot2^N,\; \sigma_s \cdot r\cdot2^N \right)
\end{equation}
where $\sigma_s$ is the noise scheduling parameter in diffusion model \cite{DiffusionDet}.

Since the overall framework follows a progressive denoising scheme, where the denoised output from preceding denoising level serves as input for subsequent denoising level. 
Therefore, each denoising level is required to handle noise within twice the expected distance. 
The final output requires the denoised points to have a maximum permissible error distance of $r$, 
which implies that the initial input noisy positive samples must fall within a distance range of $r\times2^{N}$.
In this context, the stride $r_g$ for extracting global patch embeddings is defined as $r \times 2^{N-1}$.

\subsubsection{Data Augmentation}
To balance positive and negative sample ratios, ground-truth centers undergo replicated sampling with noise superposition. 
Considering computational efficiency constraints, the maximum repetition count is limited to $K_{max}=10$ iterations to prevent computational complexity explosion during positive sample assignment. 
Given the total number of noisy points as $M$ and the number of ground-truth centers as $m_{gt}$, the replication count $k$ is calculated as:
\begin{equation}
    k=\min \left( K_{max}, \lfloor \frac{\rho \cdot M}{m_{\mathrm{gt}}} \rfloor \right) 
\end{equation}
where $\rho=0.25$ is the ratio parameter, and $\lfloor \cdot \rfloor $ represents the floor operation.

\subsubsection{MinK OTA Assignment}

During the training process, to balance the number of positive and negative samples, we employ repeated sampling of ground-truth centers. 
However, adopting the traditional SimOTA approach may lead the network predictions to overly concentrate on some objects, neglecting others of significance. 
Therefore, we have devised a progressive MinK OTA Assignment strategy to address this issue. 
\begin{algorithm} 
	\caption{MinK OTA Assignment} 
	\label{alg1} 
	\begin{algorithmic}
    
		\REQUIRE {$C_{noise}, C_{gt}, k, n, r$}
        \ENSURE {Matching Matrix: $Match_{all}$}
        
        \textcolor{Green}{\text{Step 1: Compute the matching number and distance threshold.}}
		\STATE $k_{min} \gets \max \left( 1,\lfloor k/2^{n-1} \rfloor \right) $ 
		\STATE $r_{thre} \gets r\cdot 2^{N-n+1}$
        
        \textcolor{Green}{\text{Step 2: Calculate the matching cost.}}
		\STATE $Cost \gets \mathrm{distance}\left( C_{noise},C_{gt} \right) $ 
        
        \textcolor{Green}{\text{Step 3: Under SimOTA, select $k_{min}$ samples for each target.}}
        \STATE $Match_{all} \gets \mathbf{0}$
        \FOR{ $ki=1$ to $k_{min}$}
            \STATE $Match_{ki} \gets \mathbf{0}$
            \WHILE{$\mathrm{sum}(\mathrm{sum}(Match_{ki}, \mathrm{dim}=0)=0)>0$} 
                \STATE $Ids_{unmatch} \gets \mathrm{sum}(Match_{ki}, \mathrm{dim}=0)=0$
                \STATE $Cost^\prime \gets Cost[:, Ids_{unmatch}]$ 
                \STATE \textcolor{Green}{\text{Step 3.1: For each target, pick min-cost sample.}}
                \STATE $Ids \gets \mathrm{argmin}(Cost^\prime, dim=0)$
                \STATE $Match_{ki}[Ids,Ids_{unmatch}] \gets 1$
                
                \STATE\textcolor{Green}{\text{Step 3.2: Resolve matching Conflicts}}
                \STATE $Match_{ki} \gets \mathrm{ResolveMatchingConflicts}(Match_{ki}, Cost)$
                
                \STATE\textcolor{Green}{\text{Step 3.3: Matched samples are set infinitely cost.}}
                \STATE $Cost[\mathrm{sum}(Match_{ki}, \mathrm{dim}=1)>0, :] \gets \mathrm{Inf}$
            \ENDWHILE
            \STATE $Match_{all} \gets Match_{all}\lor Match_{ki}$
        \ENDFOR
        
        \textcolor{Green}{\text{Step 4: Points assignment within the target area} }
        \STATE $Match_{d} \gets \mathbf{0}$
        \STATE $Match_{d}[Cost \leqslant r_{thre}] \gets 1$
        \STATE $Match_{d} \gets \mathrm{ResolveMatchingConflicts}(Match_{d}, Cost)$
        \STATE $Match_{all} \gets Match_{all} \lor Match_{d}$
        \RETURN $Match_{all}$

	\end{algorithmic} 
\end{algorithm}

\begin{algorithm}
 \caption{Resolve Matching Conflicts}
 \label{alg2} 
 
     \begin{algorithmic}
        \REQUIRE $ Match, Cost$
        \ENSURE $Match$
        \STATE $Ids_{multi-match}  \gets \mathrm{sum}\left( Match, \mathrm{dim}=1 \right) > 1$
        \IF{$\mathrm{any}(Ids_{multi-match})$}
            
            \STATE $Ids \gets \mathrm{argmin}(cost\left[Ids_{multi-match},: \right], dim=1)$
            \STATE $Match\left[Ids_{multi-match},: \right] \gets 0$ 
            \STATE $Match\left[Ids_{multi-match} , Ids \right] \gets 1$
        \ENDIF
        \RETURN $Match$
     \end{algorithmic}
 \end{algorithm}
Algorithm \ref{alg1} outlines the MinK OTA assignment workflow, with inputs consisting of: 
noisy sample centers $C_{noise}\in \mathbb{R}^{L_p \times 2}$, ground-truth centers $C_{gt}\in \mathbb{R}^{L_{gt} \times 2}$, positive sample replication count $k$, 
current denoising level $n$ and radius parameter $r$.
The initial phase computes two critical parameters conditioned on the current denoising level $n$:
Minimum sample number $k_{min}$ determines the least number of noisy samples required for ground-truth center matching.
Matching radius threshold $r_{thre} $ defines the maximum allowable Euclidean distance between samples and ground-truth,
which progressively tighten spatial tolerance.
Then, compute a pairwise distance matrix $Cost \in \mathbb{R}^{L_p \times L_{gt}}$ between the noisy samples and ground-truth centers. 
In order to select $k_{min}$ samples for each ground-truth center,
the loop follows SimOTA and executes $k_{min}$ times. 
Firstly, selecting the closest sample for each unmatched target. 
It is possible that multiple targets select the same sample, so the second step is to resolve conflicting matches.
Finally, the cost of the existing matching target is set to infinity to prevent repeated matching. 
If there are any remaining samples within the radius $r_{thre}$ of the target, they will be considered as positive samples and assigned to the nearest target.

 Algorithm \ref{alg2} outlines the matching conflicts resolving workflow.
There is a matching conflict when a sample is matched to multiple targets.
In this case, the sample is matched to the closer target, and the matching with other targets is reset to 0.

\begin{table*}[]
\centering
\caption{Comparison experiments on RsData dataset. The best result is marked in red and the second is in blue.}
\label{tab:rsdata}
\setlength{\tabcolsep}{2.5pt}
\resizebox{1.\linewidth}{!}{
\begin{tabular}{cccccccccccccccccccccccccc}
\hline
                                        & \multicolumn{3}{c}{}                                                                                                                    & \multicolumn{3}{c}{\textbf{Video 1}}                                                                                                    & \multicolumn{3}{c}{\textbf{Video 2}}                                                                                                    & \multicolumn{3}{c}{\textbf{Video 3}}                                                                                                    & \multicolumn{3}{c}{\textbf{Video 4}}                                                                                                    & \multicolumn{3}{c}{\textbf{Video 5}}                                                                                                    & \multicolumn{3}{c}{\textbf{Video 6}}                                                                                                    & \multicolumn{3}{c}{\textbf{Video 7}}                                                                                                    &                                      \\
                                        & \multicolumn{3}{c}{\multirow{-2}{*}{\textbf{Average}}}                                                                                  & \multicolumn{3}{c}{\textbf{(ID:3)}}                                                                                                     & \multicolumn{3}{c}{\textbf{(ID:5)}}                                                                                                     & \multicolumn{3}{c}{\textbf{(ID:2)}}                                                                                                     & \multicolumn{3}{c}{\textbf{(ID:8)}}                                                                                                     & \multicolumn{3}{c}{\textbf{(ID:10)}}                                                                                                    & \multicolumn{3}{c}{\textbf{(ID:6)}}                                                                                                     & \multicolumn{3}{c}{\textbf{(ID:9)}}                                                                                                     &                                      \\ \cline{2-25}
\multirow{-3}{*}{\textbf{Methods}}      & \textbf{Re}                          & \textbf{Pr}                          & \textbf{F1}                                               & \textbf{Re}                          & \textbf{Pr}                          & \textbf{F1}                                               & \textbf{Re}                          & \textbf{Pr}                          & \textbf{F1}                                               & \textbf{Re}                          & \textbf{Pr}                          & \textbf{F1}                                               & \textbf{Re}                          & \textbf{Pr}                          & \textbf{F1}                                               & \textbf{Re}                          & \textbf{Pr}                          & \textbf{F1}                                               & \textbf{Re}                          & \textbf{Pr}                          & \textbf{F1}                                               & \textbf{Re}                          & \textbf{Pr}                          & \textbf{F1}                                               & \multirow{-3}{*}{\textbf{FPS}}       \\ \hline
\multicolumn{1}{c|}{Vibe \cite{ViBe}}               & 0.65                                 & 0.51                                 & \multicolumn{1}{c|}{0.57}                                 & 0.61                                 & 0.34                                 & \multicolumn{1}{c|}{0.44}                                 & 0.82                                 & 0.61                                 & \multicolumn{1}{c|}{0.70}                                 & 0.68                                 & 0.59                                 & \multicolumn{1}{c|}{0.63}                                 & 0.65                                 & 0.52                                 & \multicolumn{1}{c|}{0.58}                                 & 0.72                                 & 0.65                                 & \multicolumn{1}{c|}{0.69}                                 & 0.60                                 & 0.42                                 & \multicolumn{1}{c|}{0.49}                                 & 0.45                                 & 0.44                                 & \multicolumn{1}{c|}{0.44}                                 & 0.77                                 \\
\multicolumn{1}{c|}{GoDec \cite{GoDec}}              & 0.85                                 & 0.52                                 & \multicolumn{1}{c|}{0.61}                                 & {\color[HTML]{0070C0} \textbf{0.92}} & 0.51                                 & \multicolumn{1}{c|}{0.65}                                 & 0.73                                 & 0.81                                 & \multicolumn{1}{c|}{0.77}                                 & {\color[HTML]{0070C0} \textbf{0.93}} & 0.53                                 & \multicolumn{1}{c|}{0.68}                                 & 0.72                                 & 0.38                                 & \multicolumn{1}{c|}{0.50}                                 & 0.72                                 & 0.74                                 & \multicolumn{1}{c|}{0.73}                                 & 0.81                                 & 0.42                                 & \multicolumn{1}{c|}{0.55}                                 & {\color[HTML]{FF0000} \textbf{0.93}} & 0.25                                 & \multicolumn{1}{c|}{0.39}                                 & 0.20                                 \\
\multicolumn{1}{c|}{Decolor \cite{Decolor}}            & 0.58                                 & {\color[HTML]{0070C0} \textbf{0.84}} & \multicolumn{1}{c|}{0.66}                                 & 0.24                                 & 0.92                                 & \multicolumn{1}{c|}{0.38}                                 & 0.77                                 & {\color[HTML]{0070C0} \textbf{0.88}} & \multicolumn{1}{c|}{0.82}                                 & 0.89                                 & 0.83                                 & \multicolumn{1}{c|}{0.86}                                 & 0.44                                 & {\color[HTML]{FF0000} \textbf{0.93}} & \multicolumn{1}{c|}{0.60}                                 & 0.74                                 & 0.84                                 & \multicolumn{1}{c|}{0.79}                                 & 0.71                                 & 0.80                                 & \multicolumn{1}{c|}{0.75}                                 & 0.30                                 & 0.69                                 & \multicolumn{1}{c|}{0.42}                                 & 0.12                                 \\
\multicolumn{1}{c|}{ClusterNet \cite{Clusternet}}         & 0.74                                 & 0.73                                 & \multicolumn{1}{c|}{0.73}                                 & 0.75                                 & 0.67                                 & \multicolumn{1}{c|}{0.71}                                 & 0.66                                 & 0.81                                 & \multicolumn{1}{c|}{0.72}                                 & 0.90                                 & 0.72                                 & \multicolumn{1}{c|}{0.80}                                 & 0.50                                 & 0.70                                 & \multicolumn{1}{c|}{0.58}                                 & 0.76                                 & 0.82                                 & \multicolumn{1}{c|}{0.79}                                 & 0.77                                 & 0.71                                 & \multicolumn{1}{c|}{0.74}                                 & 0.85                                 & 0.66                                 & \multicolumn{1}{c|}{0.75}                                 & 2.50                                 \\
\multicolumn{1}{c|}{DTTP \cite{DTTP}}               & 0.60                                 & 0.74                                 & \multicolumn{1}{c|}{0.65}                                 & 0.74                                 & 0.67                                 & \multicolumn{1}{c|}{0.70}                                 & 0.67                                 & 0.84                                 & \multicolumn{1}{c|}{0.74}                                 & 0.71                                 & 0.84                                 & \multicolumn{1}{c|}{0.77}                                 & 0.64                                 & 0.86                                 & \multicolumn{1}{c|}{0.73}                                 & 0.62                                 & 0.77                                 & \multicolumn{1}{c|}{0.69}                                 & 0.55                                 & 0.73                                 & \multicolumn{1}{c|}{0.62}                                 & 0.25                                 & 0.49                                 & \multicolumn{1}{c|}{0.33}                                 & 0.50                                 \\
\multicolumn{1}{c|}{AGMM \cite{AGMM}}               & 0.82                                 & 0.60                                 & \multicolumn{1}{c|}{0.68}                                 & 0.72                                 & 0.56                                 & \multicolumn{1}{c|}{0.63}                                 & 0.80                                 & 0.77                                 & \multicolumn{1}{c|}{0.79}                                 & {\color[HTML]{0070C0} \textbf{0.93}} & 0.65                                 & \multicolumn{1}{c|}{0.76}                                 & {\color[HTML]{FF0000} \textbf{0.87}} & 0.62                                 & \multicolumn{1}{c|}{0.72}                                 & 0.76                                 & 0.68                                 & \multicolumn{1}{c|}{0.72}                                 & 0.79                                 & 0.53                                 & \multicolumn{1}{c|}{0.63}                                 & {\color[HTML]{0070C0} \textbf{0.90}} & 0.37                                 & \multicolumn{1}{c|}{0.53}                                 & -                                    \\
\multicolumn{1}{c|}{E-LSD \cite{E-LSD}}              & 0.63                                 & 0.80                                 & \multicolumn{1}{c|}{0.70}                                 & 0.71                                 & 0.83                                 & \multicolumn{1}{c|}{0.77}                                 & 0.75                                 & {\color[HTML]{0070C0} \textbf{0.88}} & \multicolumn{1}{c|}{0.81}                                 & 0.64                                 & 0.67                                 & \multicolumn{1}{c|}{0.65}                                 & 0.61                                 & 0.86                                 & \multicolumn{1}{c|}{0.72}                                 & 0.57                                 & {\color[HTML]{FF0000} \textbf{0.92}} & \multicolumn{1}{c|}{0.70}                                 & 0.55                                 & 0.82                                 & \multicolumn{1}{c|}{0.66}                                 & 0.58                                 & 0.61                                 & \multicolumn{1}{c|}{0.60}                                 & 0.03                                 \\
\multicolumn{1}{c|}{D\&T \cite{DandT}}               & 0.73                                 & 0.78                                 & \multicolumn{1}{c|}{0.74}                                 & 0.71                                 & 0.91                                 & \multicolumn{1}{c|}{0.80}                                 & 0.69                                 & 0.86                                 & \multicolumn{1}{c|}{0.76}                                 & 0.84                                 & 0.84                                 & \multicolumn{1}{c|}{0.84}                                 & 0.75                                 & 0.85                                 & \multicolumn{1}{c|}{0.80}                                 & 0.63                                 & 0.82                                 & \multicolumn{1}{c|}{0.71}                                 & 0.64                                 & 0.76                                 & \multicolumn{1}{c|}{0.70}                                 & 0.83                                 & 0.43                                 & \multicolumn{1}{c|}{0.56}                                 & {\color[HTML]{FF0000} \textbf{5.56}} \\
\multicolumn{1}{c|}{B-MCMD \cite{B-MCMD}}             & 0.77                                 & 0.77                                 & \multicolumn{1}{c|}{0.76}                                 & 0.77                                 & {\color[HTML]{0070C0} \textbf{0.93}} & \multicolumn{1}{c|}{0.85}                                 & 0.76                                 & 0.86                                 & \multicolumn{1}{c|}{0.81}                                 & 0.86                                 & 0.82                                 & \multicolumn{1}{c|}{0.84}                                 & 0.71                                 & 0.77                                 & \multicolumn{1}{c|}{0.74}                                 & 0.58                                 & 0.84                                 & \multicolumn{1}{c|}{0.68}                                 & 0.70                                 & 0.74                                 & \multicolumn{1}{c|}{0.72}                                 & 0.81                                 & 0.47                                 & \multicolumn{1}{c|}{0.60}                                 & 0.02                                 \\
\multicolumn{1}{c|}{DBP \cite{DBP}}                & 0.77                                 & {\color[HTML]{FF0000} \textbf{0.85}} & \multicolumn{1}{c|}{0.81}                                 & 0.83                                 & 0.90                                 & \multicolumn{1}{c|}{0.86}                                 & 0.76                                 & {\color[HTML]{0070C0} \textbf{0.88}} & \multicolumn{1}{c|}{0.81}                                 & 0.90                                 & {\color[HTML]{FF0000} \textbf{0.88}} & \multicolumn{1}{c|}{{\color[HTML]{0070C0} \textbf{0.89}}} & 0.65                                 & 0.83                                 & \multicolumn{1}{c|}{0.73}                                 & 0.72                                 & {\color[HTML]{0070C0} \textbf{0.89}} & \multicolumn{1}{c|}{0.80}                                 & 0.73                                 & {\color[HTML]{0070C0} \textbf{0.86}} & \multicolumn{1}{c|}{0.79}                                 & 0.83                                 & 0.74                                 & \multicolumn{1}{c|}{0.78}                                 & 2.08                                 \\
\multicolumn{1}{c|}{MMB \cite{MMB}}                & 0.84                                 & {\color[HTML]{FF0000} \textbf{0.85}} & \multicolumn{1}{c|}{{\color[HTML]{0070C0} \textbf{0.84}}} & 0.83                                 & 0.84                                 & \multicolumn{1}{c|}{0.84}                                 & 0.83                                 & {\color[HTML]{FF0000} \textbf{0.89}} & \multicolumn{1}{c|}{{\color[HTML]{0070C0} \textbf{0.85}}} & {\color[HTML]{FF0000} \textbf{0.94}} & {\color[HTML]{FF0000} \textbf{0.88}} & \multicolumn{1}{c|}{{\color[HTML]{FF0000} \textbf{0.91}}} & {\color[HTML]{0070C0} \textbf{0.85}} & 0.86                                 & \multicolumn{1}{c|}{{\color[HTML]{0070C0} \textbf{0.86}}} & 0.80                                 & 0.81                                 & \multicolumn{1}{c|}{0.80}                                 & 0.78                                 & 0.85                                 & \multicolumn{1}{c|}{{\color[HTML]{0070C0} \textbf{0.81}}} & 0.83                                 & 0.73                                 & \multicolumn{1}{c|}{0.78}                                 & 2.00                                 \\
\multicolumn{1}{c|}{DSFNet \cite{DSFNet}}             & 0.85                                 & 0.83                                 & \multicolumn{1}{c|}{0.83}                                 & {\color[HTML]{FF0000} \textbf{0.95}} & 0.75                                 & \multicolumn{1}{c|}{0.84}                                 & {\color[HTML]{0070C0} \textbf{0.88}} & 0.83                                 & \multicolumn{1}{c|}{{\color[HTML]{0070C0} \textbf{0.85}}} & 0.92                                 & 0.80                                 & \multicolumn{1}{c|}{0.86}                                 & {\color[HTML]{0070C0} \textbf{0.85}} & {\color[HTML]{0070C0} \textbf{0.89}} & \multicolumn{1}{c|}{{\color[HTML]{FF0000} \textbf{0.87}}} & {\color[HTML]{0070C0} \textbf{0.85}} & 0.82                                 & \multicolumn{1}{c|}{{\color[HTML]{0070C0} \textbf{0.84}}} & 0.76                                 & {\color[HTML]{FF0000} \textbf{0.90}} & \multicolumn{1}{c|}{{\color[HTML]{FF0000} \textbf{0.82}}} & 0.71                                 & {\color[HTML]{0070C0} \textbf{0.80}} & \multicolumn{1}{c|}{0.75}                                 & {\color[HTML]{0070C0} \textbf{5.00}} \\ \hline
\multicolumn{1}{c|}{DiffMOD w/o global} & 0.85                                 & {\color[HTML]{FF0000} \textbf{0.85}} & \multicolumn{1}{c|}{{\color[HTML]{FF0000} \textbf{0.85}}} & 0.87                                 & {\color[HTML]{FF0000} \textbf{0.95}} & \multicolumn{1}{c|}{{\color[HTML]{FF0000} \textbf{0.91}}} & 0.87                                 & 0.85                                 & \multicolumn{1}{c|}{{\color[HTML]{FF0000} \textbf{0.86}}} & 0.91                                 & {\color[HTML]{0070C0} \textbf{0.85}} & \multicolumn{1}{c|}{0.88}                                 & 0.78                                 & 0.87                                 & \multicolumn{1}{c|}{0.82}                                 & 0.84                                 & 0.85                                 & \multicolumn{1}{c|}{{\color[HTML]{FF0000} \textbf{0.85}}} & 0.84                                 & 0.79                                 & \multicolumn{1}{c|}{{\color[HTML]{FF0000} \textbf{0.82}}} & 0.87                                 & {\color[HTML]{FF0000} \textbf{0.82}} & \multicolumn{1}{c|}{{\color[HTML]{FF0000} \textbf{0.85}}} & 2.31                                 \\ 
\multicolumn{1}{c|}{DiffMOD w/o TPGF}   & {\color[HTML]{FF0000} \textbf{0.89}} & 0.75                                 & \multicolumn{1}{c|}{0.81}                                 & 0.88                                 & 0.87                                 & \multicolumn{1}{c|}{0.88}                                 & 0.92                                 & 0.75                                 & \multicolumn{1}{c|}{0.83}                                 & 0.94                                 & 0.65                                 & \multicolumn{1}{c|}{0.77}                                 & 0.84                                 & 0.79                                 & \multicolumn{1}{c|}{0.82}                                 & {\color[HTML]{FF0000} \textbf{0.88}} & 0.75                                 & \multicolumn{1}{c|}{0.81}                                 & {\color[HTML]{FF0000} \textbf{0.89}} & 0.71                                 & \multicolumn{1}{c|}{0.79}                                 & 0.84                                 & 0.73                                 & \multicolumn{1}{c|}{0.78}                                 & 2.20                                 \\
\multicolumn{1}{c|}{DiffMOD}            & {\color[HTML]{0070C0} \textbf{0.87}} & 0.83                                 & \multicolumn{1}{c|}{{\color[HTML]{FF0000} \textbf{0.85}}} & 0.88                                 & 0.91                                 & \multicolumn{1}{c|}{{\color[HTML]{0070C0} \textbf{0.89}}} & {\color[HTML]{FF0000} \textbf{0.89}} & 0.83                                 & \multicolumn{1}{c|}{{\color[HTML]{FF0000} \textbf{0.86}}} & 0.92                                 & 0.83                                 & \multicolumn{1}{c|}{0.87}                                 & {\color[HTML]{0070C0} \textbf{0.85}} & 0.84                                 & \multicolumn{1}{c|}{0.85}                                 & 0.84                                 & 0.83                                 & \multicolumn{1}{c|}{{\color[HTML]{0070C0} \textbf{0.84}}} & {\color[HTML]{0070C0} \textbf{0.87}} & 0.75                                 & \multicolumn{1}{c|}{0.80}                                 & 0.86                                 & 0.72                                 & \multicolumn{1}{c|}{{\color[HTML]{0070C0} \textbf{0.79}}} & 1.68 \\ \toprule                               
\end{tabular}}
\end{table*}

\subsubsection{Loss}
The loss function comprises three components: classification loss, regression loss, and a missing loss. 
\begin{equation}
    L=\lambda_1L_{cls}+\lambda_2L_{reg}+\lambda_3L_{miss}
\end{equation}
The classification loss $L_{cls}$ adopts focal loss, while the regression loss $L_{reg}$ employs Smooth L1 loss. 
Specifically, we set $\lambda_1=2, \lambda_2=5,\lambda_3=4$ in the experiments.

During the positive sample assignment process, each ground-truth center is assigned at least $k_{min}$ sampling points, but there are still ground-truth centers where all sample distances exceed the radius threshold $r_{thre}$. 
We call this type of ground-truth center a miss target.
The miss target has a small proportion of samples assigned to it, so its contribution to the existing loss is lower than that of the target with more samples assigned to it. 
Therefore, a missing loss is incorporated into the loss function to mitigate potential sampling aggregation biases
introduced by the assignment mechanism. 

Firstly, we compute the minimum distance $dist_{min}^i$ between each target and its corresponding assigned sampling points. 
\begin{equation}
    dist_{min}^{i}=\min \left( \mathrm{dist}\left( C_{\mathrm{noise}},c_{\mathrm{gt}}^{i} \right)\right) 
\end{equation}

Subsequently, the target missing penalty term is defined as:
\begin{equation}
    L_{miss}=\frac{1}{N_{gt}}\cdot \sum_{i\in gt}{-\ln \left( \mathrm{sigmoid}\left( (1.5-\frac{dist_{\min}^{i}}{r_{thre}})\cdot \gamma _2 \right) +\epsilon \right)}
\end{equation}
Here, $r_{thre}$ is the same as that used in the progressive MinK OTA assignment. 
$\gamma_2$ controls the slope of the sigmoid function and is set to 10 in this context.
$\epsilon=1e^{-4}$ is a small constant to prevent numerical instability.
Set the distance ratio $ dist \ ratio=dist_{\min}^{i}/r_{thre}$.

\begin{figure}
    \centering
    \includegraphics[width=0.7\linewidth]{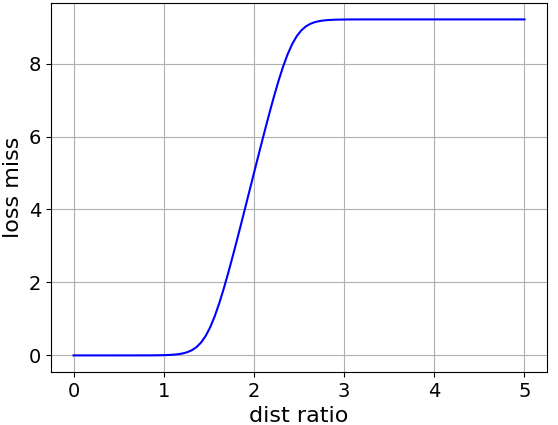}
    \caption{Plot of the function $L_{miss}$.}
    \label{fig:lossmiss}
\end{figure}

As shown in the Fig.~\ref{fig:lossmiss}, when the distance ratio is less than 1, the missing loss is close to 0. 
As the distance ratio increases, the missing loss value increases and the maximum value is $-\ln \left( \epsilon \right) $.

\section{Experiments}

\subsection{Dataset Description}
RsData is collected by DSFNet \cite{DSFNet} from from Jilin-1 satellite.
The moving vehicles in the videos were selected as the targets.
The training set and test set contain 72 and 7 satellite videos in datasets, respectively.
The training images were cropped to $512\times512$, while the testing set maintained the original resolution of $1024\times1024$. 
The video sequences contained approximately 300 frames per clip. 
The average size of target objects was $8.5\times6.7$ pixels.

\subsection{Implementation Details}
Our method is based on the Detectron2 framework \cite{detectron2} and implemented with 8 V100 GPU with 32 GB of memory. 
The backbone network in DSFNet \cite{DSFNet} is used as backbone for feature extracting.
 For temporal feature extraction, the backbone network processed 5-frame input sequences. 
 The TPGF module was trained sequentially using 3-frame inputs. 
 In training processing, the number of initial sparse noise points is set 500. 
 The model was trained using the AdamW optimizer with a batch size of 24.
 The base learning rate was set to 0.001 for a total of 15,000 iterations. 
 A warm-up strategy was applied during the first 1,000 iterations, 
 and the learning rate was reduced to $0.1\times$ of the base rate at the 12,000th iteration.

 \begin{figure*}
    \centering
    \includegraphics[width=0.95\linewidth]{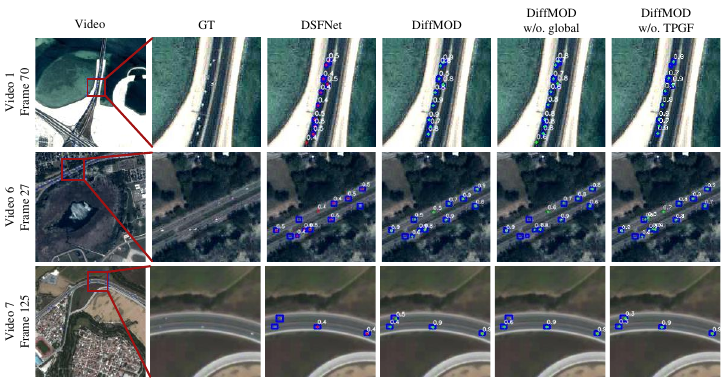}
    \caption{Visualization comparisons of detection results on satellite videos. 
    }
    \label{fig:vis}
\end{figure*}

\subsection{Evaluation on RsData Dataset}

To make a fair comparison, the evaluation procedure follows \cite{DSFNet}, \cite{viso}. 
The primary evaluation metrics contain Recall (Re), Precision (Pr), F1 score. 
As shown in Table~\ref{tab:rsdata}, model-based methods typically require longer computation times and struggle to handle noise in remote sensing , resulting in suboptimal performance. 
In contrast, learning-based approaches, particularly DSFNet \cite{DSFNet} and DiffMOD, achieve relatively higher evaluation scores.
The experimental results reveal that DSFNet \cite{DSFNet}, DiffMOD, and the variants of DiffMOD exhibit complementary strengths across different scenarios. 
For instance, in Video 1, the targets are densely distributed and visually salient. DSFNet, which relies on dense target existence probability prediction, demonstrates precise localization and effective target discrimination. 
In contrast, DiffMOD and its variants, which employ point-based modeling, tend to produce false alarms in such scenarios.
Conversely, in scenes with high inter-scene variation and low object discriminability (e.g., Video6 and Video7), point-based modeling facilitates higher-order interactions among intra-frame objects, thereby improving discrimination performance. 
While global feature embedding helps mitigate missed detections caused by the extreme sparsity of point-based representations, it introduces false positives in noisy regions such as sea surfaces and forests.
Furthermore, DiffMOD with TPGF leverages temporal information to filter out random noise across frames. 
However, this approach suffers from detection lag when dealing with blurred or newly emerging objects.

\subsection{Ablation experiments}
\subsubsection{Variants of SRAA Attention Mechanisms}
Firstly, we explore the impact of the structure of SRAA.
The ablation study results are presented as: (a) replacement of SRAA with standard self/cross-attention without spatial relations; (b) spatial-only attention without semantic affinity; (c) SRAA with both spatial relations and semantic affinity.
As shown in Table~\ref{tab:abl-SRAA}, when relying solely on spatial information, sparse point features only interact with neighboring points and global features, resulting in limited perceptual scope and significant performance degradation. 
In contrast to experiment modeling semantic relationships via self-attention and cross-attention mechanisms, 
SRAA module simultaneously exploits both spatial and semantic information. 
This integration facilitates high-order interactions among scattered point features, consequently improving both recall and precision rates in MOD tasks.

\begin{table}[]
\centering
\caption{ The impact of spatial relation in SRAA.}
\label{tab:abl-SRAA}
\begin{tabular}{cccc}
\bottomrule
                                           & \multicolumn{3}{c}{\textbf{Average}}                                                    \\ \cline{2-4} 
\multirow{-2}{*}{\textbf{Methods}}         & \textbf{Re}                 & \textbf{Pr}                 & \textbf{F1}                 \\ \hline
\multicolumn{1}{c|}{w/o. spatial relation} & 0.84                        & 0.82                        & 0.83                        \\
\multicolumn{1}{c|}{only spatial relation} & 0.78                        & \textbf{0.87}               & 0.82                        \\
\multicolumn{1}{c|}{SRAA}                  & \textbf{0.87}               & 0.83                        & \textbf{0.85}                     \\ \toprule
\end{tabular}
\end{table}

\subsubsection{Ablation Study on Progressive Training}
\begin{table}[]
\centering
\caption{Comparison of exponential and linear scheduling.}
\label{tab:abl-e}
\begin{tabular}{cccc}
\bottomrule
\multirow{2}{*}{\textbf{Methods}}         & \multicolumn{3}{c}{\textbf{Average}} \\ \cline{2-4} 
                                 & \textbf{Re}      & \textbf{Pr}      & \textbf{F1}      \\ \hline
\multicolumn{1}{c|}{uniform}     & 0.80     & 0.54    & 0.64    \\
\multicolumn{1}{c|}{exponential} & \textbf{0.87}    & \textbf{0.83}    & \textbf{0.85}  \\ \toprule
\end{tabular}
\end{table}

In the second part of our ablation study, 
we investigate how different progressive variation of matching number $k_{min}$ and distance threshold $r_{thre}$ across denoising levels affect model performance. 
Exponential scheduling and linear scheduling are compared.
With exponential scheduling, the matching number $k_{min}$ and distance threshold $r_{thre}$ are adjusted as:
\begin{equation}
\begin{split}
    k_{min} & \gets \max \left( 1,\lfloor k\cdot \small{\frac{1}{2^{n-1}}} \rfloor \right) 
    \\
    r_{thre} & \gets r\cdot 2^{N-n+1}
\end{split}
\end{equation}
With linear scheduling, the matching number $k_{min}$ and distance threshold $r_{thre}$ are adjusted as:
\begin{equation}
\begin{split}
    k_{min}^\prime &\gets \max \left( 1,\lfloor k\cdot \frac{N-n+1}{N} \rfloor \right) \\
    r_{thre}^\prime &\gets r^\prime \cdot (N-n+1)
\end{split}
\end{equation}
where $r^\prime=\frac{r\cdot 2^N}{N}$ is used in linear scheduling.

As shown in Table~\ref{tab:abl-e}, our experiments demonstrate that the linear scheduling method is considerably more difficult to train.
We observe severe imbalance in loss values across different denoising levels, where classification loss $L_{cls}$ and missing loss $L_{miss}$ in early layers decrease quickly while regression loss $L_{reg}$ remains consistently elevated. 
As progressively denoised samples reach later layers, their high dispersion and low proportion of valid samples result in slow loss convergence.
Conversely, the exponential scheduling approach yields more balanced training dynamics. 
This method establishes distinct learning focuses across denoising levels: initial layers efficiently drive samples toward their targets, while subsequent layers specialize in precisely distinguishing between different targets and achieving more refined adjustments.

\subsubsection{Comparison of Anti-clustering Strategies}

\begin{table}[]
\centering
\caption{ The impact of SimOTA and MinK OTA strategies.}
\label{tab:abl-mink}
\begin{tabular}{cccc}
\bottomrule
\multirow{2}{*}{\textbf{Methods}} & \multicolumn{3}{c}{\textbf{Average}}    \\ \cline{2-4} 
                                  & \textbf{Re} & \textbf{Pr} & \textbf{F1} \\ \hline
\multicolumn{1}{c|}{SimOTA}       & 0.73        & \textbf{0.87}        & 0.79        \\
\multicolumn{1}{c|}{MinK OTA}     & \textbf{0.87}        & 0.83        & \textbf{0.85}        \\ \toprule
\end{tabular}
\end{table}

\begin{table}[]
\centering
\caption{ The impact of missing loss.}
\label{tab:missloss}
\begin{tabular}{cccc}
\bottomrule
\multirow{2}{*}{\textbf{Methods}} & \multicolumn{3}{c}{\textbf{Average}}    \\ \cline{2-4} 
                                  & \textbf{Re} & \textbf{Pr} & \textbf{F1} \\ \hline
\multicolumn{1}{c|}{w/o. missing loss}       & 0.83        & \textbf{0.83}        & 0.83       \\
\multicolumn{1}{c|}{with missing loss}     & \textbf{0.87}  & \textbf{0.83}       & \textbf{0.85}        \\ \toprule
\end{tabular}
\end{table}

MinK OTA strategy and missing loss are designed to address the denoided point clustering. 
 We empirically validate their effects through ablation studies.
As shown in Table~\ref{tab:abl-mink} and Table~\ref{tab:missloss}, these experiments investigates the impact of two anti-clustering strategies on model performance. 

Following The SimOTA \cite{dino}  assignment strategy, each target is assigned at least one corresponding sample. 
This is generally acceptable for object detection in natural images, where the number of targets is typically much smaller than the number of detector samples and proposal boxes cover larger image regions.
However, the sample points are extremely sparse relative to the image area in DiffMOD. 
When applying SimOTA \cite{dino}, scattered points tend to cluster around salient targets to minimize classification and regression losses, ultimately leading to missed detections for other objects.
In contrast, as shown in Table~\ref{tab:abl-mink} and Table~\ref{tab:missloss}, both our designed MinK-OTA strategy and missing loss contribute to enhanced recall performance, empirically validating their capability to mitigate the scattered point clustering problem in denoising.

\subsubsection{Parameter Analysis of denoising levels $N$ and radius $r$}
As shown in Table~\ref{tab:abl-ps}, We systematically investigate the impact of different denoising levels and radius parameters on model performance. 
In remote sensing scenarios where moving objects are typically small, empirical results demonstrate that a radius of $r=4$ effectively ensures denoised points remain within the target bounding boxes. 
Here, $R= r\cdot2^{N}$ denotes the maximum acceptance radius in the first denoising level.
Due to computational resource and training time constraints, the maximum denoising level tested was limited to 4.
Our experiments reveal that when $R=128$, the acceptance fields of different objects exhibit excessive overlap, leading to competition among samples. 
As a result, the denoised points tend to cluster around the most salient objects, suppressing less prominent ones.
Conversely, with $R=32$, the augmented noise sample range becomes too restricted, deviating significantly from real-world distributions. 
This causes the network to overwhelmingly predict high-confidence outputs, resulting in a substantial increase in false alarms.

\begin{table}[]
\centering
\caption{Parameter analysis .}
\label{tab:abl-ps}
\begin{tabular}{cccccc}
\bottomrule
\multicolumn{3}{c}{\textbf{Parameters}}           & \multicolumn{3}{c}{\textbf{Average}}                                                    \\ \hline
\textbf{r} & \textbf{N} & \textbf{R}              & \textbf{Re}                 & \textbf{Pr}                 & \textbf{F1}                 \\ \hline
8          & 4          & \multicolumn{1}{c|}{128} & 0.82                        & 0.67                        & 0.74             \\
4          & 4          & \multicolumn{1}{c|}{64} & 0.87                        & \textbf{0.83}               & \textbf{0.85}            \\
4          & 3          & \multicolumn{1}{c|}{32} & \textbf{0.89}               & 0.49                        & 0.63             \\ \toprule
\end{tabular}
\end{table}

\section{Conclusion}
Existing MOD methods in remote sensing rely on dense feature extraction and object presence probability estimation, which limits information interaction and propagation across objects and temporal sequences. 
To address this issue, we represent the point-based modeling and employ progressive diffusion denoising to train the point-based MOD framework. 
Furthermore, we propose two novel modules, SRAA and TPGF, to facilitate high-order interaction and information propagation among objects and across frame, respectively. 
Experiments demonstrate that our method better adapts to scene variations and effectively detects moving targets in scenarios with weak appearance cues and strong noise interference.

However, sparse point-based representations will introduce randomness. 
While global feature enhances the detector's long-range object search capability, it also increases the false alarm rate. 
Considering the characteristics of remote sensing videos, integrating road extraction as a multi-task learning approach may yield mutual benefits. 
Additionally, the point-based modeling paradigm is well-suited for multi-object tracking tasks.

\bibliographystyle{IEEEtran}
\bibliography{reference}







\vfill

\end{document}